\documentclass[11pt]{article} % For LaTeX2e
\usepackage{jmlr2e}

\usepackage[utf8]{inputenc} % allow utf-8 input
\usepackage[T1]{fontenc}    % use 8-bit T1 fonts
\usepackage{hyperref}       % hyperlinks
\hypersetup{
    colorlinks=true,
    linkcolor=blue,
    citecolor=blue,
    urlcolor=blue,
}
\usepackage{enumitem}
\usepackage{bm}
\usepackage{url}            % simple URL typesetting
\usepackage{wrapfig,lipsum,booktabs}       % professional-quality tables
\usepackage{amsfonts}       % blackboard math symbols
\usepackage{nicefrac}       % compact symbols for 1/2, etc.
\usepackage{microtype}      % microtypography

\usepackage{caption}
\usepackage{subcaption}

\usepackage{amssymb}
\usepackage{algorithm}
\usepackage{algorithmicx}
\usepackage{framed,graphicx}
\usepackage{xcolor}
\usepackage{amsmath}
\usepackage{bigints}
\usepackage{multirow}
\usepackage{pifont}
\usepackage{xspace}
\usepackage{subcaption}

\firstpageno{1}

\begin{document}

\title{\textsc{Gecko}: An Efficient Neural Architecture Inherently Processing Sequences with Arbitrary Lengths}

\author{\name Xuezhe Ma\thanks{Equal Contribution. Correspondence to xuezhema@usc.edu and acun@meta.com} \email xuezhema@usc.edu \\
\name Shicheng Wen$*$ \email wenshich@usc.edu \\
\name Linghao Jin$*$ \email linghaoj@usc.edu \\
\addr University of Southern California  \\
\name Bilge Acun$*$ \email acun@meta.com \\
\addr Meta AI Research \\
\name Ruihang Lai$*$ \email ruihangl@cs.cmu.edu \\
\name Bohan Hou \email bohanhou@cs.cmu.edu \\
\addr Carnegie Mellon University \\
\name Will Lin \email wlsaidhi@ucsd.com \\
\name Hao Zhang \email haozhang@ucsd.edu \\
\addr University of California San Diego \\
\name Songlin Yang \email yangsl66@mit.edu \\
\addr MIT CSAIL \\
\name Ryan Lee \email ryantlee@usc.edu \\
\name Mengxi Wu \email mengxiwu@usc.edu \\
\name Jonathan May \email jonmay@usc.edu \\
\addr University of Southern California  \\
%\name Chunting Zhou\thanks{Word done while at Meta.} \email chunting.violet.zhou@gmail.com \\
\name Luke Zettlemoyer \email lsz@meta.com \\
\name Carole-Jean Wu \email carolejeanwu@meta.com \\
\addr Meta AI Research \\
}
\editor{}

\maketitle

\begin{abstract}
Designing a unified neural network to efficiently and inherently process sequential data with arbitrary lengths is a central and challenging problem in sequence modeling.
The design choices in Transformer, including quadratic complexity and weak length extrapolation, have limited their ability to scale to long sequences. 
In this work, we propose \textsc{Gecko}, a neural architecture that inherits the design of \textsc{Mega} and \textsc{Megalodon} (exponential moving average with gated attention), and further introduces multiple technical components to improve its capability to capture long range dependencies, including \emph{timestep decay normalization}, \emph{sliding chunk attention} mechanism, and \emph{adaptive working memory}.
In a controlled pretraining comparison with \textsc{Llama}2 and \textsc{Megalodon} in the scale of 7 billion parameters and 2 trillion training tokens, \textsc{Gecko} achieves better efficiency and long-context scalability.
\textsc{Gecko} reaches a training loss of \emph{1.68}, significantly outperforming \textsc{Llama}2-7B (1.75) and \textsc{Megalodon}-7B (1.70), and landing close to \textsc{Llama}2-13B (1.67). 
Notably, without relying on any context-extension techniques, \textsc{Gecko} exhibits inherent long-context processing and retrieval capabilities, stably handling sequences of up to 4 million tokens and retrieving information from contexts up to $4\times$ longer than its attention window. \\
\textbf{Code}: \url{https://github.com/XuezheMax/gecko-llm}
\end{abstract}

\begin{figure}[t]
\centering
\includegraphics[width=0.97\textwidth]{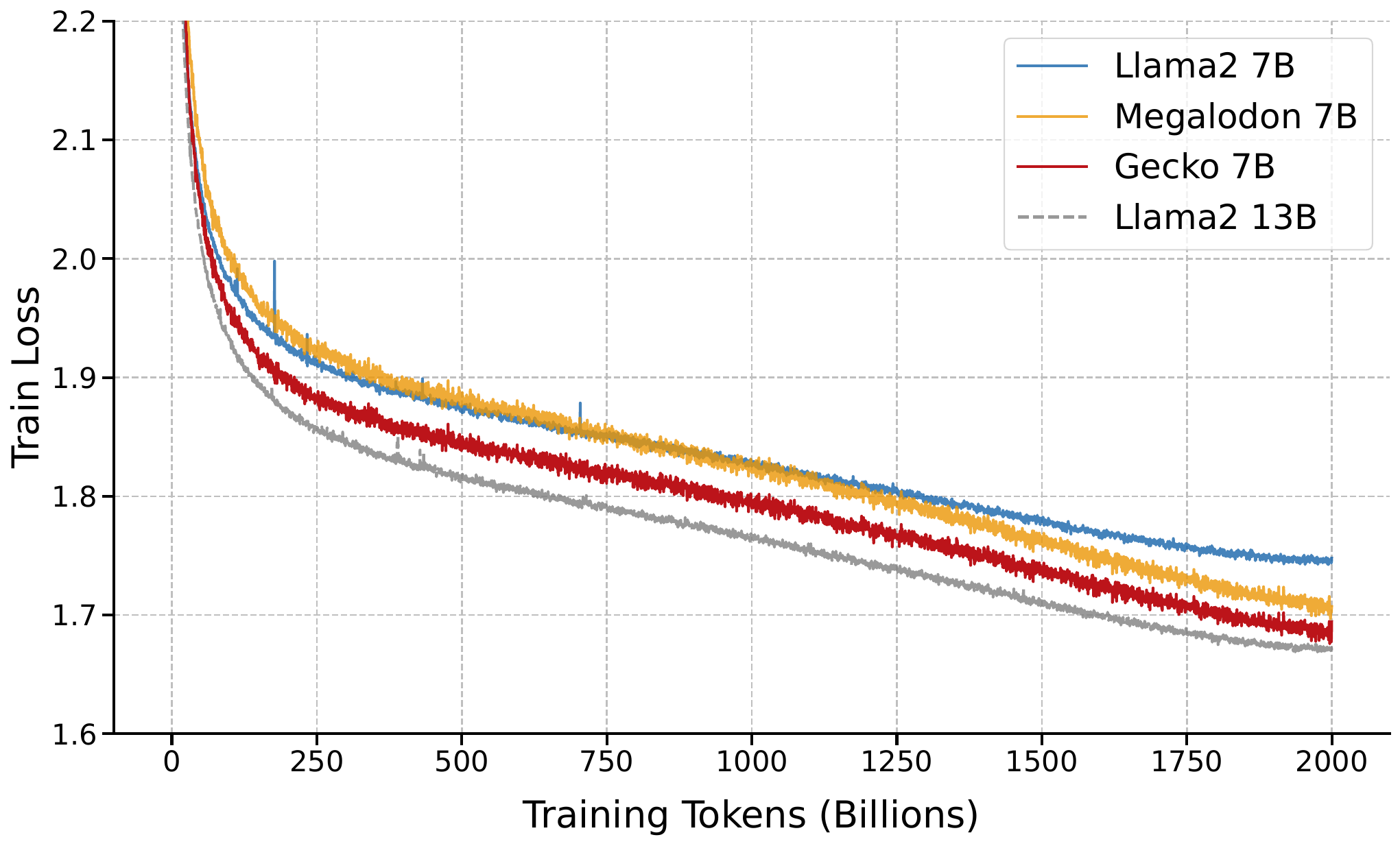}
\caption{\textbf{Negative log-likelihood} for \textsc{Gecko}-7B, \textsc{Megalodon}-7B, \textsc{Llama}2-7B and \textsc{Llama}2-13B, w.r.t processed tokens during training.
}
\label{fig:train_loss}
\vspace{-3mm}
\end{figure}

\section{Introduction}
The capability of efficient long-context modeling is crucial for the neural architecture design of next-generation large language models (LLMs). 
In many real-world applications, such as multi-turn conversation, mathematical reasoning, and video generation, large language models (LLMs) must efficiently process long sequential data, understand internal long-range dynamics, and generate coherent output.
The Transformer architecture~\citep{Vaswani+2017}, despite its remarkable capabilities, faces challenges with quadratic computational complexity and weak length extrapolation, making it inefficient for long sequence modeling~\citep{wang2024limits,zhou2024transformers,lu2025a}.

Techniques like sparse attention mechanisms~\citep{tay2020efficient,ma2021luna}, structured state space models~\citep{gu2022efficiently,poli2023hyena,gu2023mamba} and linear Transformers~\citep{katharopoulos2020transformers,mamba2,gateddeltanet} have been introduced to overcome these limitations, with the aim of improving model efficiency and performance.
However, the practical application of these methods still falls short of Transformers, particularly for in-context retrieval-oriented tasks~\citep{arora2024simple,wenrnns,gateddeltanet}.

This work aims at introducing a model architecture that is capable of efficiently and inherently processing sequences with unlimited context length, and outperforms the canonical Transformer architecture on real-world language modeling.
Moving average gated attention~(\textsc{Mega} and \textsc{Megalodon})~\citep{ma2023mega,ma2024megalodon}, which harnesses the gated attention mechanism~\citep{hua2022transformer,qiu2025gated} with the classical exponential moving average (EMA) approach~\citep{hunter1986exponentially}~(\S\ref{sec:background}), have gained significant interest due to their impressive model capacity, distributed scalability, and numerical stability.
However, the existing problems in timestep normalization~(\S\ref{subsec:timestepnorm}) and chunk-wise gated attention~(\S\ref{subsec:norm-attn}) have restricted the utilization of long-context information in \textsc{Mega}. 

In this paper, we propose \textsc{Gecko}, a novel model architecture built on the Megalodon backbone, incorporating several new techniques to further enhance the capability and efficiency of large-scale long-context pretraining and inference.
First, \textsc{Gecko} introduces a \emph{timestep decay normalization} layer that controls the influence of the current mean and variance on the cumulative statistics, keeping it at a fixed ratio (\S\ref{subsec:tsdn}).
Then, \textsc{Gecko} extends chunk-wise attention to \emph{sliding chunk attention} (SCA), which caches sliding segments on a chunk-by-chunk basis (\S\ref{subsec:sca}).
To capture long-term information that lies outside the sliding chunks, \textsc{Gecko} incorporates an \emph{adaptive working memory} component, implemented using a linear attention mechanism with a position-aware online softmax activation (\S\ref{subsec:awk}).

\begin{table}[t]
\caption{\textbf{Performance on standard academic benchmarks} of \textsc{Gecko} on two model scales (1.3B \& 7B), compared to open-source base models. We reported model size, context length (CTX) and total data tokens during model pretraining. -- indicates that the number was not reported in the original paper.}
\label{tab:benmarks}
\centering
\resizebox{\columnwidth}{!}{
\begin{tabular}{@{}lccccccccccccc@{}}
\toprule
\textbf{Model} & \textbf{Size} & \textbf{Data} & \textbf{CTX} & \textbf{MMLU} & \textbf{BoolQ} & \textbf{HellaSw} & \textbf{PIQA} & \textbf{SIQA} & \textbf{WinoG} & \textbf{Arc-e} & \textbf{Arc-c} & \textbf{NQ} & \textbf{TQA} \\
\midrule
OLMo1 & 1B & 3T & 2K & -- & 67.5 & 66.9 & 74.9 & -- & 61.4 & 55.3 & 36.5 & -- & -- \\
\textbf{\textsc{Gecko}} & 1B & 2T & 32K & 27.4 & 67.1 & 65.0 & 75.5 & 45.3 & 59.6 & 66.4 & 39.8 & 16.2 & 21.9 \\
\midrule
Mamba & 3B & 0.6T & 2K & 26.2 & 71.0 & 71.0 & 78.1 & -- & 65.9 & 68.2 & 41.7 & -- & -- \\
RWKV & 7B & 1.1T & 4K & -- & -- & 70.8 & 77.3 & -- & 68.4 & 74.9 & 46.1 & -- & -- \\
\midrule
MPT & 7B & 1T & 4K & 26.8 & 75.0 & 76.4 & 80.6 & 48.5 & 68.3 & 70.2 & 42.6 & 20.8 & 50.4 \\
%\textsc{Llama}2 Long & 7B & 2.5T & 32K & 47.8 & -- & 77.8 & 78.9 & 48.7 & 70.4 & 76.2 & 50.2 & 27.5 & 59.6 \\
Mistral & 7B & -- & 16K & 60.1 & \textbf{83.2} & \textbf{81.3} & \textbf{82.2} & 47.0 & \textbf{74.2} & 80.0 & \textbf{54.9} & 23.2 & 62.5 \\
Gemma & 8B & 6T & 8K & \textbf{64.3} & \textbf{83.2} & 81.2 & 81.2 & \textbf{51.8} & 72.3 & \textbf{81.5} & 53.2 & 23.0 & 63.4 \\
\textsc{Llama}2 & 13B & 2T & 4K & 54.8 & 81.7 & 80.7 & 80.5 & 50.3 & 72.8 & 77.3 & 49.4 & \textbf{31.2} & \textbf{65.1} \\
\midrule
\textsc{Llama}2 & 7B & 2T & 4K & 45.3 & 77.4 & 77.2 & 78.8 & 48.3 & 69.2 & 75.2 & 45.9 & 25.7 & 58.5 \\
\textsc{Megalodon} & 7B & 2T & 32K & 49.8 & 80.5 & 77.5 & 80.1 & 49.6 & 71.4 & 79.8 & 53.1 & 25.7 & 60.5 \\
\textbf{\textsc{Gecko}} & 7B & 2T & 32K & 49.4 & 81.2 & 78.1 & 80.3 & 50.2 & 73.1 & 79.3 & 53.5 & 28.5 & 61.4 \\
\bottomrule
\end{tabular}
}
\vspace{-3mm}
\end{table}

Empirically, we demonstrate the potential of \textsc{Gecko} as a general architecture for modeling long sequences, by evaluating its performance across multiple scales of language modeling, as well as downstream domain-specific tasks.
Through a direct comparison by controlling for data and compute, \textsc{Gecko}-7B significantly outperforms the state-of-the-art variant of Transformer and \textsc{Mega} used to train \textsc{Llama2}-7B~\citep{touvron2023llama} and \textsc{Megalodon}-7B~\citep{ma2024megalodon} on both training perplexity (Figure~\ref{fig:train_loss}) and across downstream benchmarks (Table~\ref{tab:benmarks}).
Evaluation on long-context modeling, including perplexity in various context lengths up to 4 million and long-context QA tasks in Scrolls~\citep{parisotto2020stabilizing} prove \textsc{Gecko}'s ability to model sequences of unlimited length.
To further assess \textsc{Gecko}’s retrieval capability, we evaluate it on two standard needle-in-a-haystack (NIAH) benchmarks: \textbf{passkey retrieval} and \textbf{vanilla NIAH}~\citep{niah2023,hsieh2024ruler}.
Across both tasks, \textsc{Gecko} retrieves information from sequences that are roughly four times longer than its attention context.
%On a smaller 1.3B model scale, \textsc{Gecko} outperforms OLMo1 .

\vspace{-3mm}
\section{Background: Backbone Architectures in \textsc{Mega}}
\label{sec:background}
\vspace{-1mm}
In this section, we set the notation and discuss existing problems of \textsc{Mega} and \textsc{Megalodon}~\citep{ma2023mega,ma2024megalodon}.
We use $\boldsymbol{X} = \{\mathbf{x}_1, \mathbf{x}_2, \ldots, \mathbf{x}_n\} \in \mathbb{R}^{n\times d}$ and $\boldsymbol{Y} =\{\mathbf{y}_1, \mathbf{y}_2, \ldots, \mathbf{y}_n\} \in \mathbb{R}^{n\times d}$ to denote the input and output sequences with length $n$, and assume that the representations of the input and output sequences have the same dimension $d$. 

\subsection{CEMA: Complex Multi-dimensional Damped EMA}
\label{subsec:cema}
\textsc{Mega} embeds an EMA component into the calculation of the attention matrix to incorporate inductive biases across the timestep dimension, while \textsc{Megalodon} further extends to work over the complex number system $\mathbb{C}$ to improve capacity.
Concretely, CEMA first expands each dimension of the input sequence $\boldsymbol{X}$ individually into $h$ dimensions via an expansion matrix $\boldsymbol{\beta} \in \mathbb{R}^{d\times h}$, then applies damped EMA to the $h$-dimensional hidden space. 
Formally, for each dimension $j \in \{1, 2, \ldots, d\}$:
\begin{align}
\label{eq:cema}
\mathbf{u}^{(j)}_t & = \boldsymbol{\beta}_j \mathbf{x}_{t,j} \nonumber \\
\mathbf{h}^{(j)}_t & = \boldsymbol{\alpha}_j \odot \mathbf{u}^{(j)}_t + (1 - \boldsymbol{\alpha}_j \odot \boldsymbol{\delta}_j) (\cos{\theta_j + i \sin{\theta_j}}) \odot \mathbf{h}^{(j)}_{t-1} \\
\mathbf{y}_{t,j} & = \mathrm{Re}(\boldsymbol{\eta}^T_j \mathbf{h}^{(j)}_t ) \nonumber 
\end{align}
where $\odot$ is the element-wise product and $\mathbf{u}^{(j)}_t \in \mathbb{R}^{h}$ is the expanded $h$-dimensional vector for the $j$-th dimension at timestep $t$. 
$\boldsymbol{\alpha} \in (0, 1)^{d\times h}$, $\boldsymbol{\delta} \in (0, 1)^{d\times h}$ are the decaying and damping factors, respectively. $\mathbf{h}^{(j)}_t \in \mathbb{C}^{h}$ is the complex hidden state for the $j$-th dimension at timestep $t$. 
$\boldsymbol{\eta} \in \mathbb{C}^{d\times h}$ is the projection matrix to map the $h$-dimensional hidden state back to $1$-dimensional output $\mathbf{y}_{t,j} \in \mathbb{R} $.
$\theta_j \in \mathbb{R}^{h}, \,\, j \in \{1, 2, \ldots, d\}$ are the $h$ arguments uniformly spaced over the period $2\pi$:
\begin{equation}
\theta_{j,k} = \frac{2\pi k}{h} \omega_j, \quad \forall k \in \{1, 2, \ldots, h\}
\end{equation}
where the learnable parameter $\omega \in \mathbb{R}^{d}$ depicts the $d$ base angles.\footnote{See \textsc{Megalodon}~\citep{ma2024megalodon} for more details.}
%By decaying the absolute value of each $h_t$, CEMA preserves the decaying structure in kernel weights, which is a key principle to the success of convolutional models on long sequence modeling~\citep{li2023makes}.

\subsection{Timestep Normalization}
\label{subsec:timestepnorm}
\textsc{Megalodon} proposed Timestep Normalization, which extends Group Normalization~\citep{wu2018group} to the auto-regressive case by computing the cumulative mean and variance.
Similar to Group Normalization, each input vector $\mathbf{x}_t$ is split into $k$ groups along the feature dimension with $d_g = d/k$ elements per group.
We use $\mu_t$ and $\sigma^2_t$ to denote the mean and variance of the first group of the input vector at timestep $t \in \{1, 2, \ldots, n \}$.
Then, the cumulative mean ($m_t$) and variance ($v_t$) are the average of the means and variances from the previous $t$ timesteps:
\begin{align}
\mu_t = \frac{1}{d_g}\sum\limits_{j=1}^{d_g} x_{i,j}, & \qquad \sigma^2_t = \frac{1}{d_g}\sum\limits_{j=1}^{d_g} (x_{i,j} - \mu_t)^2 \label{eq:mean_var} \\
m_t = \frac{1}{t}\sum\limits_{i = 1}^{t} \mu_t, & \qquad v_t = \frac{1}{t}\sum\limits_{i = 1}^{t}\sigma^2_t \label{eq:tsn}
\end{align}

\paragraph{Existing Problem.}
However, a clear drawback of Timestep Normalization is that the influence of the current mean and variance ($\mu_t$ and $\sigma_t^2$) on the cumulative statistics ($m_t$ and $v_t$) decreases monotonically as the timestep $t$ increases.
It limits the ability to inherently scale \textsc{Megalodon} to long sequences.

\subsection{Chunk-wise Normalized Gated Attention}
\label{subsec:norm-attn}

To improve efficiency and stability, \textsc{Megalodon} simply splits the sequences of queries, keys and values in (\ref{eq:q}-\ref{eq:v}) into chunks of length $c$, and embeds the CEMA component into the calculation of normalized gated attention.
Formally, 
\begin{align}
\boldsymbol{X}' & = \mathrm{CEMA}(\boldsymbol{X}) \qquad & \qquad \in \mathbb{R}^{n\times d} \\
\boldsymbol{Z} & = \boldsymbol{X}' W_z + b_z, \quad \boldsymbol{Z}' = \frac{\boldsymbol{Z}}{\| \boldsymbol{Z}\|} & \qquad \in \mathbb{R}^{n\times z} \label{eq:nz} \\
\boldsymbol{Q} & = \boldsymbol{\kappa}_q \odot \boldsymbol{Z}' + \boldsymbol{\nu}_q \qquad & \qquad \in \mathbb{R}^{n\times z} \label{eq:q} \\
\boldsymbol{K} & = \boldsymbol{\kappa}_k \odot \boldsymbol{Z}' + \boldsymbol{\nu}_k \qquad & \qquad \in \mathbb{R}^{n\times z} \label{eq:k} \\
\boldsymbol{V} & = \phi_{\mathrm{silu}}(\boldsymbol{X} W_v + b_v) \qquad & \quad \qquad \in \mathbb{R}^{n\times v} \label{eq:v} 
\end{align} 
where $\boldsymbol{\kappa}_q$, $\boldsymbol{\nu}_q$, $\boldsymbol{\kappa}_k$, $\boldsymbol{\nu}_k \in \mathbb{R}^{z}$ are the learnable scalars and offsets of queries and keys.
The attention is individually applied to each chunk, yielding linear complexity $O(kc^2)=O(nc)$:
\begin{align}\label{eq:norm-attention}
\boldsymbol{O}_s & = f_{\mathrm{softmax}} \left(\boldsymbol{Q}_s{\boldsymbol{K}_s}^{T} \right) \boldsymbol{V}_s \quad & \qquad \qquad \in \mathbb{R}^{c\times v}
\end{align}
where $s \in \{1, 2, \ldots, n/c \}$ is the index of chunks with chunk size $c$. 
$\boldsymbol{Q}_s$ and $\boldsymbol{K}_s$ and $\boldsymbol{V}_s$ are the queries, keys and values in the $s$-th chunk.

\paragraph{Existing Problems.}
Technically, the CEMA sub-layer in \textsc{Megalodon} helps capture local contextual information near each token, mitigating the problem of losing contextual information beyond chunk boundaries in the chunk-wise attention.
Despite its impressive successes, \textsc{Megalodon} still suffers from at least two limitations: 
(i) losses near the context boundaries of chunk-wise attention (both preceding and following a chunk) increase substantially, due to the limited expressiveness of the CEMA sub-layer (Figure~\ref{fig:position-nll}); 
and (ii) \textsc{Megalodon} is unable to effectively capture historical information outside the chunk-wise attention context, resulting in suboptimal performance on in-context retrieval–oriented tasks.

\section{\textsc{Gecko}}
\label{sec:gecko}

%To address the aforementioned problems of \textsc{Megalodon}, in this section we describe the novel technical advancements of \textsc{Gecko}.

\subsection{Timestep Decay Normalization}
\label{subsec:tsdn}
In order to control the influence of the current mean and variance of the cumulative statistics in Timestep Normalization \eqref{eq:tsn}, \textsc{Gecko} incorporates decaying mechanism into the calculation of cumulative statistics. 
Formally, we introduce two hyper-parameters, $\beta_1, \beta_2 \in [0, 1)$, to fix the ratio of current mean and variance, respectively:
\begin{align}
m_t = \beta_1 m_{t-1} + (1 - \beta_1) \mu_t, & \qquad v_t = \beta_2 v_{t-1} + (1 - \beta_2)\sigma^2_t \label{eq:tsdn}
\end{align}
where $\mu_t$ and $\sigma^2_t$ are the current mean and variance defined in \eqref{eq:mean_var}.
Directly inspired from the first and second order momentums in Adam~\citep{kingma2015adam}, we apply the initialization bias correction terms to both $m_t$ and $v_t$:
\begin{align}
m'_t = \frac{\mu_t}{1-\beta_1^t}, & \qquad v'_t = \frac{v_{t}}{1-\beta_2^t} \label{eq:tsdn2}
\end{align}
In practice, good default settings are $\beta_1=0.999, \beta_2=0.9999$.
Similar to Timestep Normalization in \textsc{Megalodon}, we provide hardware-friendly implementation of Timestep Decay Normalization on modern hardware (GPU).

\begin{figure*}[t]
\centering
\includegraphics[width=1\linewidth]{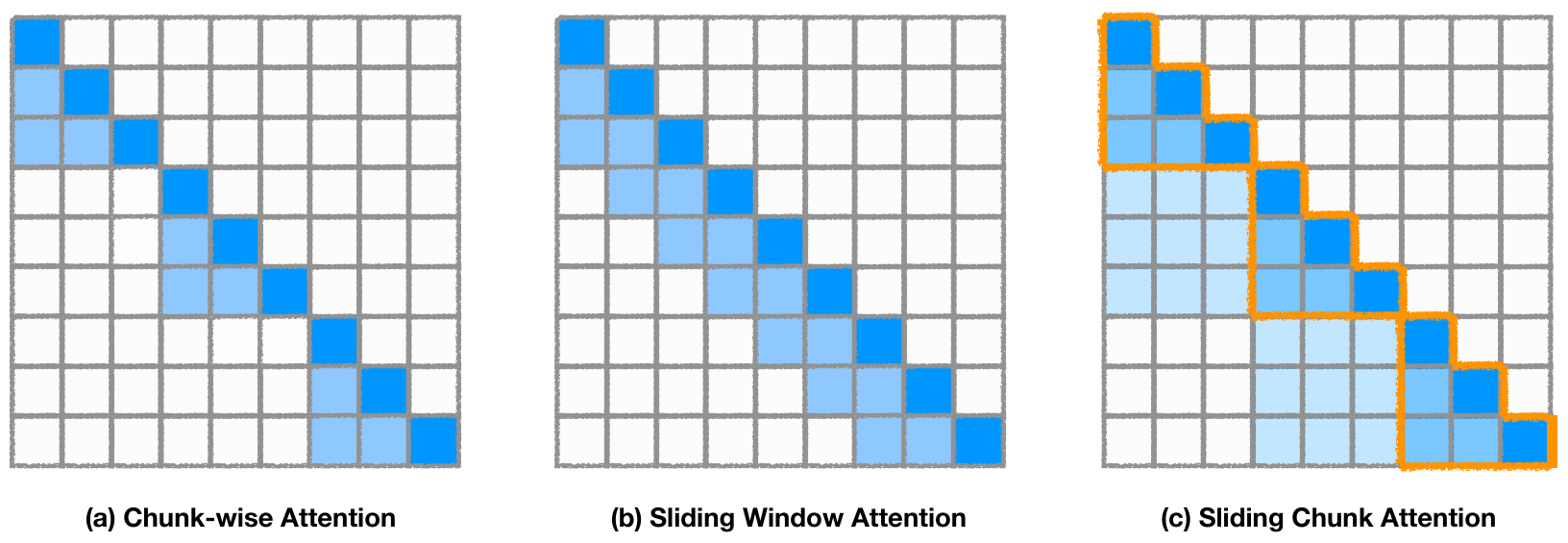}
\caption{Comparison of various sparse attention patterns. (a) Chunk-wise Attention: attention is restricted to separated chunks; (b) Sliding Window Attention: attention is restricted to fixed-size windows; (c) Sliding Chunk Attention: attention is assigned to both current and previous chunks.}
\label{fig:attn_design}
\vspace{-2mm}
\end{figure*}

\subsection{Sliding Chunk Attention}
\label{subsec:sca}
To mitigate the issues of truncated context at chunk boundaries inherent in chunk-wise attention, a straightforward approach is to replace it with sliding window attention~(SWA)~\citep{beltagy2020longformer}, which restricts each token's attention context to a fixed-size local window. 
Although SWA has lower theoretical complexity, its practical implementation often relies on many small matrix multiplications—effectively one per token—due to the per-token shifted attention window.
These operations are inefficient on modern hardware devices such as GPUs, which are designed and optimized for large, contiguous matrix computations rather than numerous small ones.

Motivated by the recurrent segment mechanism in Transformer-XL~\citep{dai2019transformer} and the block sliding window in LongFormer~\citep{beltagy2020longformer}, we extend sliding window attention by incorporating chunked contexts, yielding the \emph{sliding chunk attention} (SCA) mechanism.
Similar to chunk-wise attention, SCA partitions the input sequence $\boldsymbol{X} = \{\mathbf{x}_1, \mathbf{x}_2, \ldots, \mathbf{x}_n\} \in \mathbb{R}^{n\times d}$ into non-overlapping chunks of fixed length $c$. Specifically, the $s$-th chunk is defined as
$\boldsymbol{X}_s = \{\mathbf{x}_{(s-1)\times c + 1}, \mathbf{x}_{(s-1)\times c + 2}, \ldots, \mathbf{x}_{s\times c}\} \in \mathbb{R}^{c\times d}$,
where $s \in \{1, \ldots, n/c\}$, with query, key, and value chunks constructed in the same manner.
For ease of exposition, we denote $\boldsymbol{X}_s = \{\mathbf{x}_{s,1}, \mathbf{x}_{s,2}, \ldots, \mathbf{x}_{s,c}\}$, where $\mathbf{x}_{s,i} = \mathbf{x}_{(s-1)\times c + i}, \,\, \forall i \in \{1, \ldots, c\}$.
We use analogous notation for $\boldsymbol{Q}_s$, $\boldsymbol{K}_s$ and $\boldsymbol{V}_s$.
For each chunk, attention is computed jointly over the current chunk and the preceding chunk, enabling efficient local context propagation across chunk boundaries:
\begin{align}\label{eq:sca}
\boldsymbol{O}_s & = f_{\mathrm{softmax}} \left(\boldsymbol{Q}_s [{\boldsymbol{K}_{s-1}}, {\boldsymbol{K}_s}]^{T} \right) [\boldsymbol{V}_{s-1}, \boldsymbol{V}_s] & \qquad \in \mathbb{R}^{c\times v}
\end{align}
where $[{\boldsymbol{K}_{s-1}}, {\boldsymbol{K}_s}] \in \mathbb{R}^{2c\times d}$ denotes the concatenation of keys vectors from the previous and current chunks; similar definition for $[{\boldsymbol{V}_{s-1}}, {\boldsymbol{V}_s}] \in \mathbb{R}^{2c\times v}$. 
Despite having theoretical complexity comparable to SWA, SCA performs matrix multiplications at the chunk level rather than per token, resulting in a more efficient implementation on modern GPU hardware.
Figure~\ref{fig:attn_design} illustrates the three different sparse attention patterns.

The keys and values from the previous chunk ($\boldsymbol{K}_{s-1}, \boldsymbol{V}_{s-1}$) effectively serve as a lossless short-term memory, retaining the most recent contextual information.
However, both SCA and SWA remain limited in capturing long-range dependencies due to their constrained receptive fields, even with deep layer stacking~\citep{xiaoefficient}.
Consequently, there remains an urgent need for an effective and efficient long-term memory mechanism to process context information beyond sliding chunks.

\paragraph{Efficient Context Parallelism for SCA. }
Importantly, sliding chunk attention is well suited for efficient context parallelism in distributed pretraining.
As with CEMA and Timestep Decay Normalization, SCA requires communication only of the previous key–value chunks among devices within each context-parallel group.
Through asynchronous communication, these transfers can be overlapped with the computation of other submodules in the same block, effectively hiding communication latency and reducing parallelization overhead.

\subsection{Adaptive Working Memory}
\label{subsec:awk}
Fixed-size compressive memory mechanism promises stable and efficient computations tailored to specific contexts for extremely long sequences~\citep{kanerva1988sparse,munkhdalai2019metalearned}.
Recent work, such as \textsc{Flash}~\citep{hua2022transformer} and Infini-attention~\citep{munkhdalai2024leave}, proposed to implement compressive memory as a learnable module with linear attention mechanism~\citep{katharopoulos2020transformers,schlag2020learning}.

\paragraph{Compressive Memory with Linear Attention.}
With Infini-attention as an example, with linear attention as compressive memory, information that moves beyond attention window is recurrently compressed into a fixed-size state. 
Formally, $\mathcal{M}_s \in \mathbb{R}^{d \times v}$ denote the memory state for the $s$-th chunk.
Once moving to the next chunk of information, compressive memory is updated by applying linear attention with delta rule~\citep{widrow1988adaptive,schlag2021linear}:
\begin{align}
\mathcal{M}_s & = \mathcal{M}_{s-1} + \phi(\boldsymbol{K}_s)^T \left( \boldsymbol{V}_s - \frac{\psi(\boldsymbol{K}_s) \mathcal{M}_{s-1}}{\psi(\boldsymbol{K}_s) \boldsymbol{\tau}_{s-1}} \right) & \in \mathbb{R}^{d \times v} \label{eq:deltanet} \\
\boldsymbol{\tau}_{s} & = \sum\limits_{t=1}^{s}\sum\limits_{i=1}^{c}\phi\left(\mathbf{k}_{t,i}\right) = \boldsymbol{\tau}_{s-1} + \sum\limits_{i=1}^{c} \phi\left(\mathbf{k}_{s,i}\right) & \in \mathbb{R}^{d\times 1} \label{eq:tau} \\
\boldsymbol{O}_s & = \frac{\psi(\boldsymbol{Q}_s) \mathcal{M}_{s-1}}{\psi(\boldsymbol{Q}_s) \boldsymbol{\tau}_{s-1}} & \in \mathbb{R}^{d \times v} \label{eq:linear_out}
\end{align}
where $\boldsymbol{\tau}_{s} \in \mathbb{R}^{d}$ is the normalization term, $\phi(\cdot)$ and $\psi(\cdot)$ are the nonlinear feature kernels for keys and queries, respectively.
Commonly used featur1349415e kernels are element-wise nonlinear activation functions, such as SiLU~\citep{ramachandran2017swish}.
The output $\boldsymbol{O}_s$ is computed by retrieving information from the memory $\mathcal{M}_{s-1}$ with the query $\boldsymbol{Q}_s$ in \eqref{eq:linear_out}.
The delta update rule dynamically erases less redundant information associated with current keys and values and memory content, to make space for new ones.

Technically, the compressive memory $\mathcal{M}_s$ in linear attention maintains a bounded-capacity storage. 
As tokens move beyond the attention context, their information is incrementally accumulated in the memory.
When modeling very long sequences, however, the constrained memory capacity inevitably leads to \emph{memory collisions}, degrading information fidelity.
To address this limitation, prior work introduces gated mechanisms that act as decay factors, enabling the model to selectively forget historical information~\citep{mamba2,gateddeltanet}.
While these approaches achieve strong empirical performance, the deliberate forgetting of historical information conflicts with the design motivation of \textsc{Gecko}, which aims to retain long-term memory.
\vspace{-1mm}
\paragraph{Adaptive Working Memory (AWM) with Position-aware Online Softmax Kernel.}
To effectively compress long-term information into fixed-size memory while avoid forgetting historical one, \textsc{Gecko} introduces the \emph{position-aware online softmax kernel} to linear attention. 
For clarity of presentation, we denote the current and accumulative denominators of the softmax function in chunk $s$ as 
\begin{align}
w_s & = \sum\limits_{i=1}^{c} \exp(\mathbf{k}_{s,i}) & \qquad \qquad \qquad \qquad \in \mathbb{R}^{d \times 1} \\
z_s & = \sum\limits_{t=1}^{s}\sum\limits_{i=1}^{c} \exp(\mathbf{k}_{t,i}) = z_{s-1} + w_s & \quad \in \mathbb{R}^{d \times 1}    
\end{align}
Note that for each input vector $\mathbf{k}_{t,i} \in \mathbb{R}^d$, $\exp(\cdot)$ is the element-wise exponential function.
Then, we define the local and global feature kernels of keys with online softmax:
\begin{align}
\phi(\mathbf{k}_{s,t}) & \stackrel{\Delta}{=} f_{\mathrm{softmax}}(\mathbf{k}_{s,t}) = \frac{\exp(\mathbf{k}_{s,t})}{w_s} & \in \mathbb{R}^{d \times 1} \\
\phi_s(\mathbf{k}_{s,t}) & \stackrel{\Delta}{=} f_{\mathrm{softmax}}(\mathbf{k}_{s,t}; \mathbf{k}_{<s,t}) = \frac{\exp(\mathbf{k}_{s,t})}{z_s} = \frac{w_s}{z_s} \odot \phi(\mathbf{k}_{s,t}) & \in \mathbb{R}^{d \times 1} \label{eq:phi}
\end{align} 
For both $\phi(\cdot), \,\, \phi_s(\cdot)$, the softmax functions perform normalization over the timestep dimension of the sequence.
In contrast to the element-wise feature kernels ($\phi(\cdot)$ and $\psi(\cdot)$ in \eqref{eq:deltanet}), \emph{the position-aware kernel $\phi_s(\mathbf{k}_{s,t})$ integrates all historical information into the denominator of the online softmax function \eqref{eq:phi}}.

The normalization term $\psi(\cdot)\boldsymbol{\tau}_{s}$ in \eqref{eq:deltanet} and \eqref{eq:linear_out} might result in numerical instabilities, and has been removed it from recent work~\citep{yang2024gated,yang2024parallelizing}. 
Note that with the online softmax $\phi_s(\cdot)$ \eqref{eq:phi}), we have $\boldsymbol{\tau}_{s} = 1, \forall s = 1, 2, \ldots$. 
\textsc{Gecko} further proposes to use the softmax function along the feature dimension as the query feature kernel $\psi(\cdot)$:
\begin{align}
\psi(\mathbf{q}) & \stackrel{\Delta}{=} f_{\mathrm{softmax}}(\mathbf{q}) = \exp(\mathbf{q}) / \sum\limits_{j=1}^{d} \exp(\mathbf{q}_j) & \qquad \qquad \qquad \in \mathbb{R}^{d \times 1} \label{eq:psi}
\end{align}
where softmax normalization in $\psi(\cdot)$ is performed along the feature dimension of a query vector $\mathbf{q} \in \mathbb{R}^{d\times 1}$. 
Together with $\phi_s(\cdot)$, \textsc{Gecko} essentially eliminates the normalization term  by constraining it to be a constant of value one: 
\begin{equation}
\psi(\boldsymbol{Q}_s) \boldsymbol{\tau}_{s-1} = 1, \,\, \forall s = 1, 2, \ldots, 
\end{equation}
%which significantly facilitates the derivations of the update rule of the adaptive working memory in \textsc{Gecko}.
Then, by incorporating $\phi(\cdot), \,\, \phi_s(\cdot)$ into the memory update rule in \eqref{eq:deltanet}, we have
\begin{align}
\mathcal{M}_s & = \frac{z_{s-1}}{z_{s}} \odot \mathcal{M}_{s-1} + \frac{w_{s}}{z_{s}} \odot \phi(\mathbf{K}_s)^T \left( \mathbf{V}_s - \psi(\mathbf{K}_s) \mathcal{M}_{s-1} \right) & \quad \in \mathbb{R}^{d \times v} \label{eq:awk}
\end{align}
with the initial memory $\mathcal{M}_0 = 0$ (detailed derivations in Appendix~\ref{appendix:awk}). 
%$\odot$ is the element-wise product.

\begin{figure*}[t]
    \centering
    \begin{subfigure}{0.48\linewidth}
        \centering
        \includegraphics[width=\linewidth]{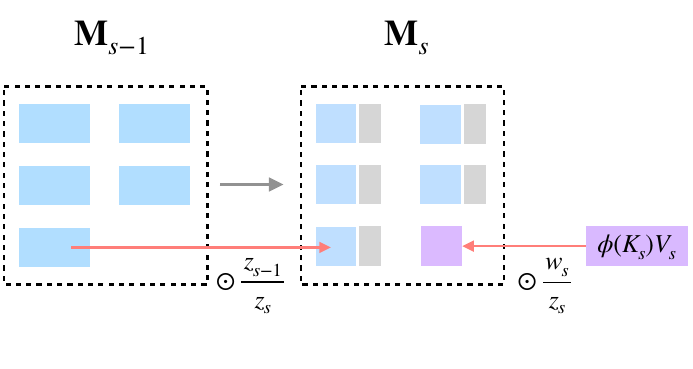}
        \caption{Adaptive Working Memory}
        \label{fig:memory}
    \end{subfigure}
    \hfill
    \begin{subfigure}{0.48\linewidth}
        \centering
        \includegraphics[width=\linewidth]{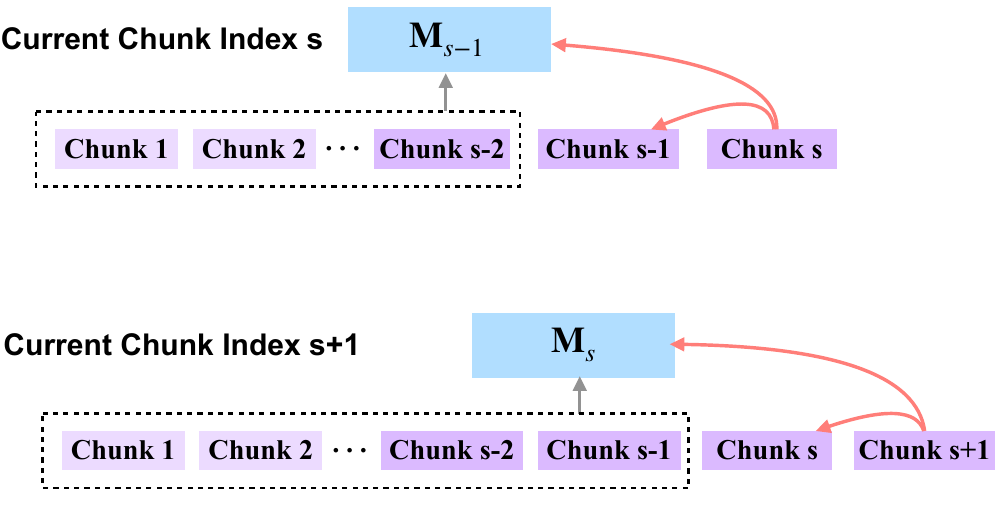}
        \caption{Attention from timestep $s$ to $s+1$.}
        \label{fig:attention_chunk}
    \end{subfigure}
    \caption{Adaptive working memory in \textsc{Gecko}. (a) illustrates how the memory is updated by compressing information from all previous and current chunks. (b) depicts how \textsc{Gecko} stores contextual information in short- and long-term memories in different components.}
    \label{fig:memory_dynamics}
\vspace{-3mm}
\end{figure*}

\paragraph{Integrating Memory with Sliding Chunk Attention.}
To integrate the short-term memory in SCA with the long-term adaptive working memory (AWM) in \eqref{eq:awk}, we need to shift the chunk index to the left by one so that the adaptive working memory captures information moving beyond the sliding context in SCA.
Concretely, denoting $\mathcal{M}_{s}' = \mathcal{M}_{s-1}$, we have the new memory updated rule with shifted chunk index $s$:
\begin{align}
\mathcal{M}_{s}' & = \frac{z_{s-2}}{z_{s-1}} \odot \mathcal{M}_{s-1}' + \phi_{s-1}(\mathbf{K}_{s-1})^T \left( \mathbf{V}_{s-1} - \psi(\mathbf{K}_{s-1}) \mathcal{M}_{s-1}' \right)
\end{align}
with the initial memory $\mathcal{M}_1' = 0$.
Figure~\ref{fig:memory_dynamics} depicts the compressive memory update mechanism (\ref{fig:memory}) as well as the coordination between short- and long-term memory components across different \textsc{Gecko} modules (\ref{fig:attention_chunk}) in maintaining contextual information.

\paragraph{Neural Parameterization.}
Inspired by Infiniti-Attention, we reuse intermediate representations from SCA computation.
In particular, we reuse the normalized shared representation $\boldsymbol{Z}'$ \eqref{eq:nz} and the value vectors $\boldsymbol{V}$ \eqref{eq:v}.
For memory computation, we introduce two additional parameter sets that act as learnable scalars and offsets to derive new keys and queries:
\begin{align}
\boldsymbol{Q}_m & = \boldsymbol{\eta}_q \odot \boldsymbol{Z}' + \boldsymbol{\rho}_q \qquad & \qquad \in \mathbb{R}^{n\times z} \label{eq:mq} \\
\boldsymbol{K}_m & = \boldsymbol{\eta}_k \odot \boldsymbol{Z}' + \boldsymbol{\rho}_k \qquad & \qquad \in \mathbb{R}^{n\times z} \label{eq:mk}
\end{align}
where $\boldsymbol{\eta}_q$, $\boldsymbol{\rho}_q$, $\boldsymbol{\eta}_k$, $\boldsymbol{\rho}_k \in \mathbb{R}^{z}$ are the learnable parameters for $\boldsymbol{Q}_m$ and $\boldsymbol{K}_m$.
Finally, we retrieve content $\mathbf{O}_m$ from the memory $\mathcal{M}_{s-1}'$ by using the query $\boldsymbol{Q}_m$:
\begin{align}
\mathbf{O}_m & = \psi(\mathbf{Q}_{m}) \mathcal{M}_{s-1}' & \qquad \qquad \in \mathbb{R}^{c \times v} 
\end{align}
The AWM output $\mathbf{O}_m$ is directly added to the SCA output in \eqref{eq:sca} for the final output.

\paragraph{Relation to Previous Work.}
The channel-wise ratio vector $\frac{z_{s-1}}{z_{s}}$ is reminiscent of the "writing strength" and  forget gate in Gated DeltaNet~\citep{gateddeltanet} and Kimi Delta Attention~\citep{team2025kimi}. 
However, AWM differs from them in two fundamental aspects.
First, AWM leverages $\frac{z_{s-1}}{z_s}$ and $\frac{w_s}{z_s}$ to globally compress information from current and all previous chunks into memory, rather than discarding historical information through forgetting.
Second, Gated DeltaNet and Kimi Delta Attention are proposed as drop-in replacements for causal full attention and perform hidden-state updates at every token step.
By contrast, AWM operates at the chunk level, updating memory on a chunk-by-chunk basis.

\section{Experiments}
\label{sec:exp}

%To assess \textsc{Gecko}’s capacity, we scale \textsc{Gecko} to two different scales --- 1-billion- and 7-billion–parameter models, and conduct large-scale language-model pretraining on 2 trillion tokens for both of them.

\subsection{LLM Pretraining Setup}

\paragraph{\textsc{Gecko}-1B vs. OLMo1.} \label{subsec:gecko1b} 
For a controlled comparison study, we configure the \textsc{Gecko}-1B model to closely follow the architectural hyperparameters of OLMo-1B (\citealt{Groeneveld2023OLMo}, July 2024 release). Both models have 16 blocks with a feature dimension of $d = 2048$. 
We pretrained Gecko 1B with 2 trillion tokens on the same Dolma v1.7 dataset~\citep{dolma} that was used to train OLMo-1B.
%ensuring that both models are trained on an identical corpus of approximately 1.7 trillion tokens. 
We also use OLMo-1B's tokenizer with a vocabulary size of 50{,}304. Training is performed using the AdamW optimizer~\citep{loshchilov2019decoupled} with $\beta_1 = 0.9$, $\beta_2 = 0.95$, $\epsilon = 1\mathrm{e}{-8}$, and a peak learning rate of $4\mathrm{e}{-4}$. Cosine decay schedule is used, along with 2{,}000 warmup steps, weight decay of 0.05, gradient clipping of 1.0, and no dropout. Additional details and experiments are provided in Appendix~\ref{appendix:experiments:gecko1b}.

\paragraph{\textsc{Gecko}-7B vs. \textsc{Llama}2.} 
In designing the \textsc{Gecko}-7B model, we closely follow the architectural hyperparameters of \textsc{Llama}2-7B and \textsc{Megalodon}-7B to enable a fair comparison. The model comprises 32 blocks with a feature dimension of $d = 4096$. As in \textsc{Megalodon}-7B, we employ SwiGLU~\citep{shazeer2020glu} in the feed-forward layers and rotary positional embeddings (RoPE; \citealt{su2021roformer}).
\textsc{Llama}2 is pretrained with a 4096-token context window, and \textsc{Megalodon}-7B uses a chunk size of $c=4096$ for its chunk-wise attention.
To maintain comparable effective context length, we set the chunk size of \textsc{Gecko}’s sliding chunk attention to $c=2048$, yielding the same 4096-token attention context.

Consistent with \citet{ma2024megalodon}, we use the same mixture of publicly available data as \textsc{Llama}2, ensuring that all models are trained on an identical corpus of 2 trillion tokens. We also adopt the \textsc{Llama}2 tokenizer with a 32K vocabulary.
Training is performed using AdamW with $\beta_1 = 0.9$, $\beta_2 = 0.95$, $\epsilon = 1\mathrm{e}{-8}$, and a peak learning rate of $3.5\mathrm{e}{-4}$. We use a cosine decay schedule with 2{,}500 warmup steps, along with weight decay of 0.1, gradient clipping of 1.0, and no dropout.
Following \textsc{Megalodon}-7B, we pretrain with a 32K context window and a global batch size of 4M tokens. Training is distributed across 256 NVIDIA H100 GPUs (16K tokens per GPU), with data parallelism of 128, chunk parallelism of 2.

\paragraph{Pretraining Results.}
Figure~\ref{fig:train_loss} presents the negative log-likelihood (NLL) curves for \textsc{Gecko}-7B in comparison with \textsc{Llama}2-7B, \textsc{Megalodon}-7B, and \textsc{Llama}2-13B over the course of training.
Across all stages, \textsc{Gecko}-7B attains noticeably lower NLL than \textsc{Llama}2-7B and \textsc{Megalodon}-7B for the same number of processed tokens, demonstrating superior data efficiency.
At convergence, \textsc{Gecko} achieves a training loss of 1.68, outperforming \textsc{Llama}2-7B (1.75) and \textsc{Megalodon}-7B (1.70), and nearly matching \textsc{Llama}2-13B (1.67).

\subsection{Short-Context Evaluation on Academic Benchmarks}
We evaluate \textsc{Gecko}-7B against \textsc{Llama}2 and \textsc{Megalodon}-7B on standard short-context academic benchmarks ($<4$K tokens), following the evaluation protocol of \textsc{Llama}2~\citep{touvron2023llama}.
The benchmarks span four categories:
\begin{itemize}[leftmargin=*]
    \item \textbf{Commonsense Reasoning} (0-shot): HellaSwag~\citep{zellers2019hellaswag}, PIQA~\citep{bisk2020piqa}, SIQA~\citep{sap2019socialiqa}, WinoGrande~\citep{sakaguchi2021winogrande}, ARC-e and -c~\citep{clark2018think}.
    \item \textbf{World Knowledge} (5-shot): NaturalQuestions (NQ, \citet{kwiatkowski2019natural}) and TriviaQA (TQA, \citet{joshi2017triviaqa}).
    \item \textbf{Reading Comprehension} (0-shot): BoolQ~\citep{clark2019boolq}.
    \item \textbf{Aggregated Evaluation} (5-shot): MMLU~\citep{hendrycks2020measuring}.
\end{itemize}
\vspace{-1mm}
Table~\ref{tab:benmarks} reports results for \textsc{Gecko}-7B, \textsc{Megalodon}-7B, and \textsc{Llama}2 models, along with other open-source baselines such as MPT~\citep{mpt7b}, RWKV~\citep{peng2023rwkv}, Mamba~\citep{gu2023mamba}, Mistral~\citep{jiang2023mistral} and Gemma~\citep{team2024gemma}.
Under identical pretraining conditions (2T tokens), \textsc{Gecko}-7B consistently outperforms both \textsc{Llama}2-7B and \textsc{Megalodon}-7B across all benchmarks. On several tasks, its performance is comparable to or even exceeds that of \textsc{Llama}2-13B.
We emphasize that Mistral-7B and Gemma-8B were pretrained on significantly larger datasets, so comparison with \textsc{Gecko}-7B is not entirely apples-to-apples.

\subsection{Long-Context Evaluation}
\begin{figure*}[t]
    \centering
    \begin{subfigure}{0.48\linewidth}
        \centering
        \includegraphics[width=\linewidth]{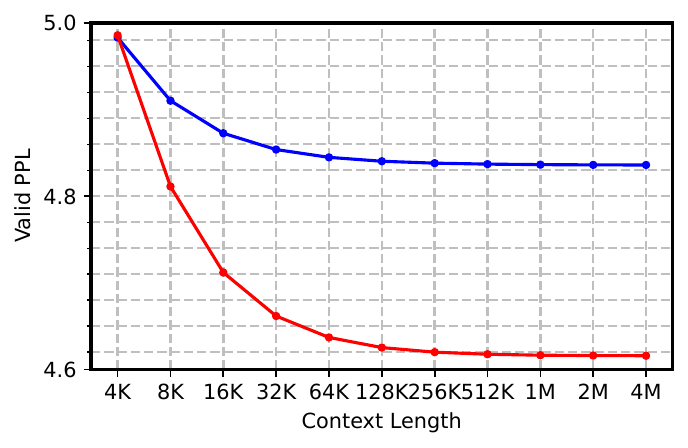}
        \caption{PPL in various context lengths.}
        \label{fig:ctx-ppl}
    \end{subfigure}
    \hfill
    \begin{subfigure}{0.48\linewidth}
        \centering
        \includegraphics[width=\linewidth]{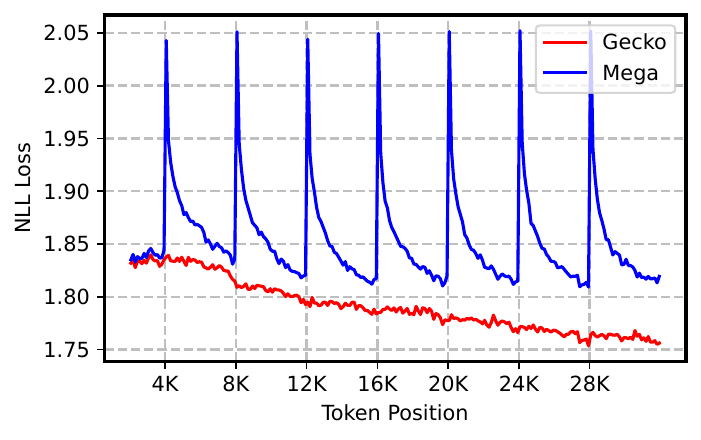}
        \caption{NLL Loss by token positions.}
        \label{fig:position-nll}
    \end{subfigure}
    \caption{PPL/NLL over long sequences. (a) shows the perplexity (PPL) in various context lengths. (b) plots averaged negative log-likelihood (NLL) broken down by token positions.}
    \label{fig:context-ppl}
    \vspace{-3mm}
\end{figure*}

\paragraph{Validation loss over Long Sequences}
To assess \textsc{Gecko}’s ability to intrinsically exploit very long contexts for next-token prediction, we evaluate validation perplexity over a range of context lengths.
Following \citet{ma2024megalodon}, we construct a validation set of 1,500 books, each containing sequences of at least 4M tokens.
Figure~\ref{fig:ctx-ppl} presents the perplexity (PPL) of \textsc{Gecko} and \textsc{Megalodon} on this dataset as the context length increases from 4K to 4M tokens.
Across all settings, \textsc{Gecko} exhibits markedly better context utilization than \textsc{Megalodon}.
Moreover, the consistent decrease in PPL with longer contexts demonstrates \textsc{Gecko}’s inherent ability—without relying on any explicit context-extension mechanisms—to model extremely long sequences.

To further analyze the effectiveness and robustness of \textsc{Gecko} in long-context modeling, we examine the average negative log-likelihood as a function of token position (up to 32K), shown in Figure~\ref{fig:position-nll}.
Because chunk-wise attention discards contextual information across chunk boundaries, \textsc{Megalodon} suffers from a pronounced increase in loss near these boundaries.
By contrast, \textsc{Gecko} exhibits steadily decreasing loss with increasing context length, highlighting its improved utilization of long-range context.

\begin{figure*}[t]
    \centering
    \begin{subfigure}{0.48\linewidth}
        \centering
        \includegraphics[width=\linewidth]{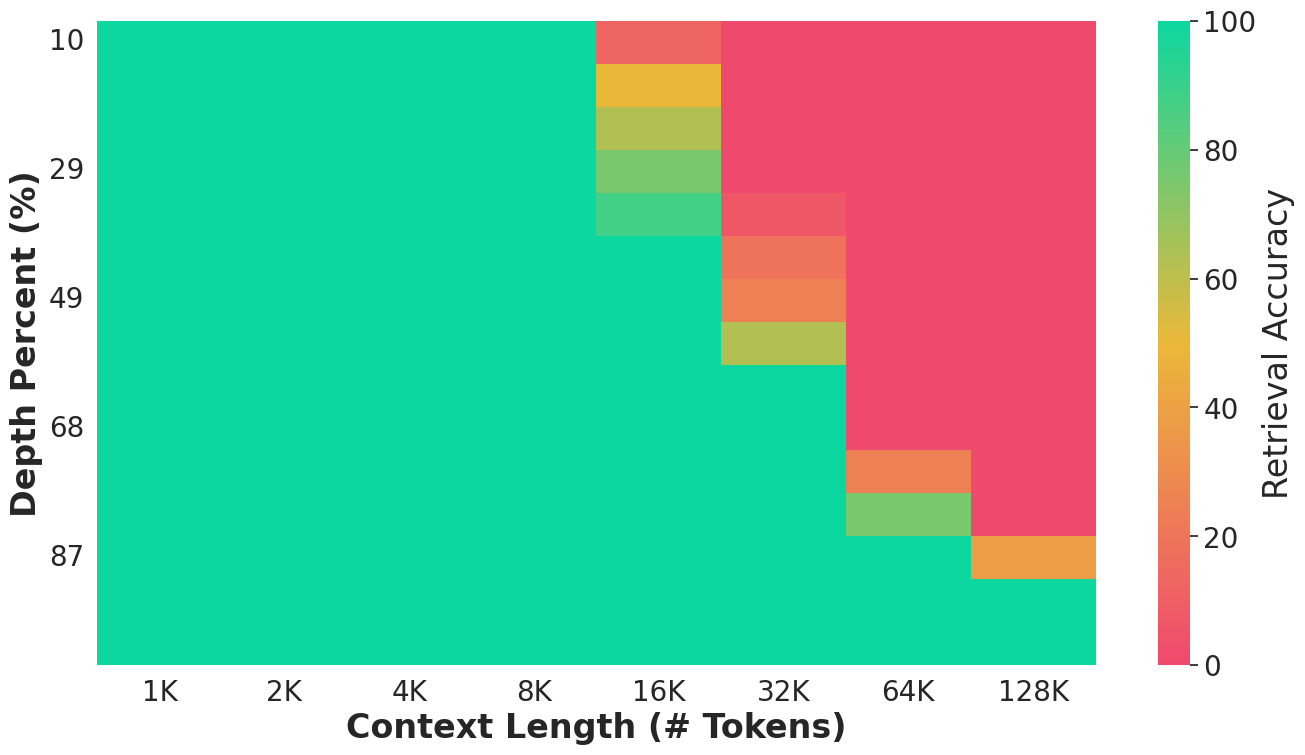}
        \caption{Passkey}
        \label{fig:passkey}
    \end{subfigure}
    \hfill
    \begin{subfigure}{0.48\linewidth}
        \centering
        \includegraphics[width=\linewidth]{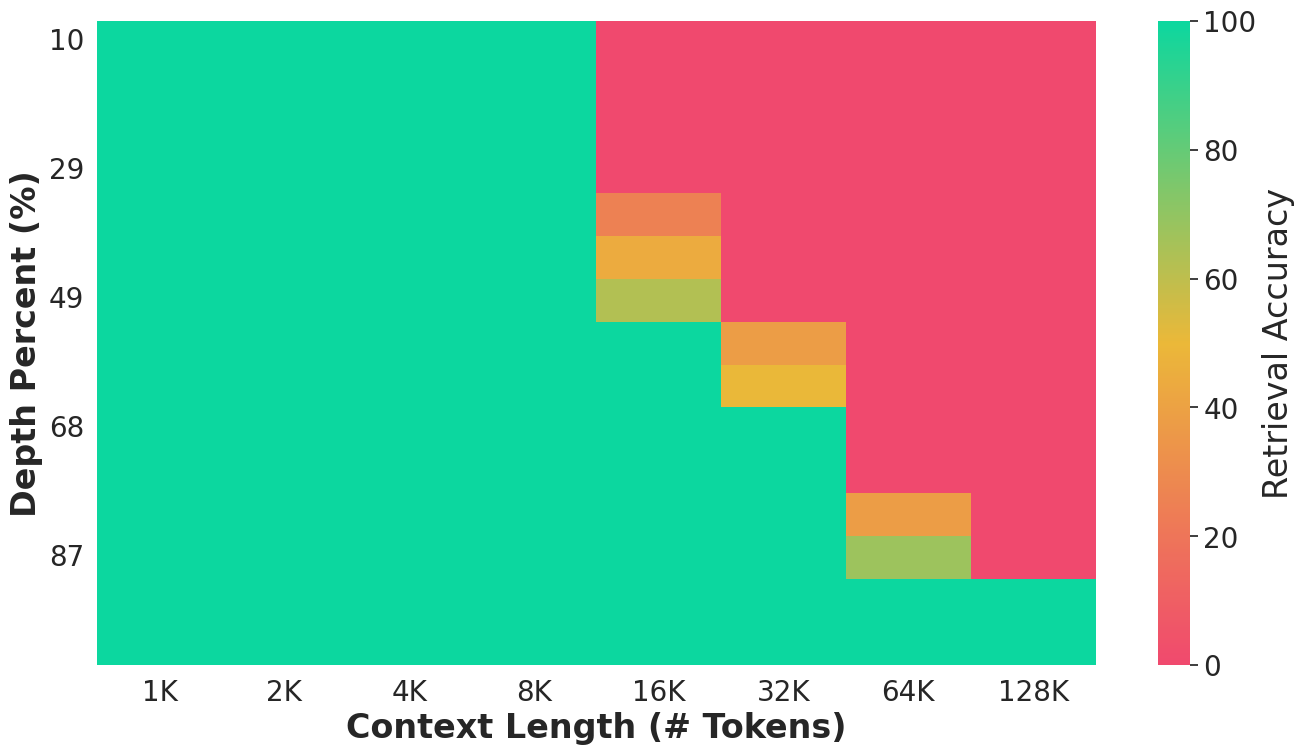}
        \caption{NIAH}
        \label{fig:niah}
    \end{subfigure}
    \caption{Evaluate Gecko-7B Long context on Passkey Retrieval and Needle in A Haystack.}
    \label{fig:retrieval}
\vspace{-3mm}
\end{figure*}

\paragraph{Long-Range Retrieval-Oriented Tasks.}
To assess \textsc{Gecko}’s information retrieval capability, we evaluate it on a single needle-in-a-haystack (NIAH) task~\citep{niah2023}.
Following the Ruler benchmark suite~\citep{hsieh2024ruler}, Figure~\ref{fig:retrieval} presents results under two haystack settings: (i) passkey retrieval with repeated synthetic context, and (ii) NIAH with real-world essay context.
For both settings, we report performance across varying needle depths.

With only 4K attention context, \textsc{Gecko} achieves near-perfect performance up to 16K sequence length, under the passkey-retrieval setting.
Under the more different NIAH setting, the performance on 16K sequence length is slightly worse than that of passkey retrieval, but still obtains $100\%$ performance for 8K length across all depth levels. 
These results demonstrate the strong inherent retrieval capability of \textsc{Gecko} without relying on any context-extension methods.

\begin{wraptable}{r}{0.55\textwidth}
\vspace{-5mm}
\captionof{table}{\textbf{Results on Scrolls}. $^*$ \textsc{Llama}2-L~\citep{xiong2023effective} continually trains \textsc{Llama}2 on 500B tokens for length extension.}
\label{tab:scrolls}
\centering
\begin{tabular}[b]{@{}lccc@{}}
\toprule
\textbf{Model} & \textbf{NaQA} & \textbf{Qasper}  & \textbf{QMSum} \\
\midrule
Xgen & 17.4 & 20.5 & ~~6.8 \\
MPT & 18.8 & 24.7 & ~~8.8 \\
Yarn & 20.9 & 26.2 & 11.4 \\
\midrule
\textsc{Llama}2 & 18.8 & 19.8 & 10.1 \\
\textsc{Llama}2-L$^*$ &  23.5 & 28.3 & 14.5 \\
\textsc{Megalodon} & 23.9 & 28.0 & 13.1 \\
\midrule
\textsc{Gecko} &  \textbf{27.3} & \textbf{34.2} & \textbf{15.8} \\
\bottomrule
\end{tabular}
\vspace{-3mm}
\end{wraptable}

\paragraph{Long-Context QA tasks in Scrolls}
We further evaluate \textsc{Gecko} on long-context open-book question answering tasks from the Scrolls benchmark~\citep{shaham-etal-2022-scrolls}, including NarrativeQA~\citep{kocisky-etal-2018-narrativeqa}, Qasper~\citep{dasigi-etal-2021-dataset} and QMSum~\citep{zhong-etal-2021-qmsum}.
Following the evaluation protocol of \citet{xiong2023effective}, we report 0-shot F1 on NarrativeQA, 2-shot F1 on Qasper, and 1-shot geometric ROUGE (the geometric mean of ROUGE-1, ROUGE-2, and ROUGE-L) on QMSum, using a unified prompt format across tasks.\footnote{Prompt format: \{\textup{CONTEXT}\} Q: \{\textup{QUESTION}\} A:}
Table~\ref{tab:scrolls} compares \textsc{Gecko}-7B and \textsc{Megalodon}-7B against other open-source long-context models of similar scale, including Xgen-7B-8K~\citep{nijkamp2023xgen7b}, MPT-7B-8K~\citep{mpt7b}, YaRN-7B-128k~\citep{peng2024yarn}, \textsc{Llama}2-7B-4K~\citep{touvron2023llama}, and \textsc{Llama}2-7B-32K (\textsc{Llama}2-L, \citet{xiong2023effective}).
Across all three benchmarks, \textsc{Gecko}-7B consistently achieves the strongest performance.
Notably, \textsc{Llama}2-7B-32K extends the base \textsc{Llama}2-7B model’s context window from 4K to 32K through continued pretraining on an additional 500B long-context tokens.

\section{Conclusion}
We present \textsc{Gecko}, a novel architecture built upon the Megalodon backbone that introduces several key innovations to enhance the efficiency and effectiveness of large-scale long-context pretraining and inference.
Through the integration of timestep decay normalization, sliding chunk attention, and adaptive working memory, \textsc{Gecko} inherently supports efficient processing of sequences with effectively unbounded context length, surpassing the canonical Transformer in real-world language modeling.
Compared directly with \textsc{Llama}2 and \textsc{Megalodon}, \textsc{Gecko} delivers consistent improvements in training perplexity and downstream benchmark performance.
Importantly, without any explicit context-extension mechanisms, \textsc{Gecko} achieves robust long-context processing and retrieval, stably handling sequences of up to 4 million tokens and retrieving information from contexts $4\times$ longer than its nominal attention context.

\section*{Acknowledgments}
We thank Chunting Zhou and Xiaomeng Yang for their helpful feedback and discussion during this work.

\bibliography{ref}

\newpage

\section*{Appendix: \textsc{Gecko}: An Efficient Neural Architecture Inherently Processing Sequences with Arbitrary Lengths}
\appendix

\section{Adaptive Working Memory}
\label{appendix:awk}
\begin{align}
\mathcal{M}_s & = \sum\limits_{c=1}^{s-1} \phi_s(\mathbf{K}_c)^T\mathbf{V}_c + \phi_s(\mathbf{K}_s)^T\mathbf{V}_s & \in \mathbb{R}^{d \times v} \\
 & = \sum\limits_{c=1}^{s-1} \left(\frac{\phi_{s}(\mathbf{K}_c)}{\phi_{s-1}(\mathbf{K}_c)} \odot \phi_{s-1}(\mathbf{K}_c) \right)^T \mathbf{V}_c + \phi_s(\mathbf{K}_s)^T\mathbf{V}_s & \in \mathbb{R}^{d \times v} \\
 & = \sum\limits_{c=1}^{s-1} \left(\frac{z_{s-1}}{z_{s}} \odot \phi_{s-1}(\mathbf{K}_c) \right)^T \mathbf{V}_c + \phi_s(\mathbf{K}_s)^T\mathbf{V}_s & \in \mathbb{R}^{d \times v} \\
 & = \frac{z_{s-1}}{z_{s}} \odot \mathcal{M}_{s-1} + \phi_s(\mathbf{K}_s)^T\mathbf{V}_s & \in \mathbb{R}^{d \times v} \\
 & = \frac{z_{s-1}}{z_{s}} \odot \mathcal{M}_{s-1} + \frac{w_{s}}{z_{s}} \odot \phi(\mathbf{K}_s)^T\mathbf{V}_s & \in \mathbb{R}^{d \times v}
\end{align}
Applying delta rule, we have:
\begin{align}
\mathcal{M}_s & = \frac{z_{s-1}}{z_{s}} \odot \mathcal{M}_{s-1} + \phi_s(\mathbf{K}_s)^T \left( \mathbf{V}_s - \psi(\mathbf{K}_s) \mathcal{M}_{s-1} \right) & \quad \in \mathbb{R}^{d \times v} 
\end{align}
where $\mathcal{M}_0 = 0$.
The output of chunk $s$ is:
\begin{align}
\mathbf{O}_s & = \psi(\mathbf{Q}_s) \mathcal{M}_{s-1} & \qquad \qquad \qquad \qquad \qquad \qquad \qquad \qquad \in \mathbb{R}^{c \times v} 
\end{align}

\paragraph{Incorporating with Sliding Chunk.}
$\mathcal{M}_{s}' = \mathcal{M}_{s-1}$, via incorporating with sliding chunks, we have:
\begin{align}
\mathcal{M}_{s}' & = \frac{z_{s-2}}{z_{s-1}} \odot \mathcal{M}_{s-1}' + \phi_{s-1}(\mathbf{K}_{s-1})^T \left( \mathbf{V}_{s-1} - \psi(\mathbf{K}_{s-1}) \mathcal{M}_{s-1}' \right) & \quad \in \mathbb{R}^{d \times v}
\end{align}
where $\mathcal{M}_1' = 0$.
The output of chunk $s$ is:
\begin{align}
\mathbf{O}_s & = \psi(\mathbf{Q}_s) \mathcal{M}_{s-1}' & \qquad \qquad \qquad \qquad \qquad \qquad \qquad \qquad \in \mathbb{R}^{c \times v} 
\end{align}

\section{Implementation Details}

\subsection{Efficient Fused CUDA Operators Implementation}
\label{appendix:cuda}
The CEMA recurrence poses a challenge for efficient training on GPUs due to its sequential dependency. Standard PyTorch implementations suffer from high memory I/O and low core utilization. To address this, we implement a highly optimized CUDA Kernel that reformulates the recurrence as a parallel associative scan, utilizing custom warp-level primitives and hardware-aware memory layouts.

\subsubsection{Parallel Associative Scan Formulation}
To enable parallelization of Equation \ref{eq:cema} via the prefix scan algorithm, we map this linear recurrence to a complex affine algebra. For a given feature dimension $j$, we define the multiplicative term $q_t \in \mathbb{C}$ and the additive term $p_t \in \mathbb{C}$ as:
\begin{equation}
q_t = (1 - \alpha_j \odot \delta_j)(\cos \theta_j + i \sin \theta_j), \quad p_t = \alpha_j \odot u_t^{(j)}
\end{equation}
The recurrence allows us to define a binary associative operator $\otimes$ acting on tuples $S_t = (q_t, p_t)$. The operation $(q_b, p_b) \otimes (q_a, p_a)$ combines a current state $b$ with a previous state $a$:
\begin{equation}
(q_b, p_b) \otimes (q_a, p_a) = (q_b \cdot q_a, \ q_b \cdot p_a + p_b)
\end{equation}
This formulation reduces the sequence modeling complexity from linear span $O(L)$ to logarithmic span $O(\log L)$ on parallel hardware.
\paragraph{Zero-Cost Boundary Handling.} To handle document boundaries inherent in variable-length processing, we avoid control flow divergence by embedding boundary logic directly into the algebra. Using a pre-computed mask $M_t \in \{0, 1\}$ (where 0 indicates a document reset), we compute the effective multiplicative term as $q_t^{\text{eff}} = q_t \cdot M_t$. When $M_t=0$, the state naturally resets ($h_t = p_t$), allowing uniform instruction execution across irregular boundaries.

\subsubsection{Kernel Architecture and Thread Mapping}
Our kernel grid minimizes kernel launch overhead and maximizes instruction-level parallelism by exploiting the specific dimensions of the hidden state $H \in \mathbb{C}^{B \times D \times N \times L}$.

\begin{itemize}
    \item \textbf{Feature-Major Grid:} We map the feature dimension $D$ to the CUDA Grid, where each thread block processes a tile of features.
    \item \textbf{Warp-Level Parallelism:} We adopt a "one-warp-per-feature" strategy. Each warp (32 threads) is assigned a unique feature index $d$ and processes the sequence $L$. Crucially, $B$ and $N$ dimensions are handled sequentially within the warp via loops. This Parameter Reuse strategy enables time-invariant parameters ($p, q, \gamma$) to be loaded once into Shared Memory/L1 Cache and reused $B \times N$ times, significantly increasing arithmetic intensity.
    \item \textbf{Chunked Warp Scan:} We partition sequence $L$ into chunks of size 32. Inside each chunk, threads utilize warp shuffle intrinsics to perform register-level cooperative scans. This avoids the latency of round-tripping to Shared Memory for intra-chunk dependencies.
\end{itemize}

\subsubsection{Memory Hierarchy and Vectorization}
To saturate HBM, we implement strictly coalesced access patterns:

\begin{itemize}
    \item \textbf{Vectorized \texttt{float4} Storage:} Since CEMA operates on complex numbers, we utilize \texttt{float4} data types to load the affine tuple $(q, p)$. Each \texttt{float4} stores 128 bits representing $[\text{Re}(q), \text{Im}(q), \text{Re}(p), \text{Im}(p)]$. This vectorization ensures that every global memory transaction is 128-bit aligned and coalesced.
    \item \textbf{Shared Memory Tiling:} Input chunks $X$ and parameters are prefetched into Shared Memory buffers. This hides global memory latency behind the arithmetic computations of the prefix scan.
\end{itemize}

\subsubsection{Memory-Efficient Backward Pass}
Storing the full hidden state $H \in \mathbb{C}^{B \times D \times N \times L}$ for gradient computation is prohibitive for long-context training. We employ a chunk-level rematerialization strategy:

\begin{itemize}
    \item \textbf{Forward Pass:} We only store the chunk boundaries (the accumulated state at the end of every 32-element chunk) to global memory.
    \item \textbf{Backward Pass:} The kernel reconstructs intermediate $h_t$ values on-the-fly by re-running the forward prefix scan from the nearest stored chunk boundary, immediately followed by a backward adjoint scan.
\end{itemize}

This approach reduces the activation memory footprint by a factor of $32\times$, enabling the scaling of GECKO to long sequences without memory overflow.

\section{Experimental Details}
\label{appendix:experiments}

\begin{table}[h]
\caption{\textbf{Performance on standard academic benchmarks} of \textsc{Gecko}-1B compared to OLMo1-1B at matched training checkpoints (approximately 2T tokens). We report model size, context length (CTX) and total data tokens during model pretraining.}
\label{tab:benmarks_appendix}
\centering
\resizebox{\columnwidth}{!}{
\begin{tabular}{@{}lccccccccc@{}}
\toprule
\textbf{Model} & \textbf{Size} & \textbf{Data} & \textbf{CTX} & \textbf{BoolQ} & \textbf{HellaSw} & \textbf{PIQA} & \textbf{WinoG} & \textbf{Arc-e} & \textbf{Arc-c} \\
\midrule
OLMo1 & 1B & 2T & 2K & 63.9 & 49.5 & 73.1  & 
\textbf{62.4} & 55.1 & 35.8  \\
\textbf{\textsc{Gecko}} & 1B & 2T & 32K & \textbf{67.1} & \textbf{65.0} & \textbf{75.5} & 59.6 & \textbf{66.4} & \textbf{39.8} \\
\bottomrule
\end{tabular}
}
\vspace{-3mm}
\end{table}

\subsection{Experiment on \textsc{Gecko}-1B} 
\label{appendix:experiments:gecko1b}
As reported in Table \ref{tab:benmarks} and Section \ref{subsec:gecko1b}, we conducted a controlled comparison between a 1B version of Gecko and OLMo1-1B (0724). Due to resource constraints, we did not pretrain Gecko-1B up to 3T tokens as OLMo1-1B did, instead training up to 2T tokens using the Dolma v1.7 dataset. For as close a comparison as possible, during training we also matched OLMo1-1B's global batch size, i.e. number of tokens per optimizer step. Even with less training, we see comparable results against the published OLMo1 1B benchmark performance and better results for PIQA, ARC-Easy, and ARC-Challenge. For an additional point of comparison, we report in Table \ref{tab:benmarks_appendix} Gecko-1B at 2024B tokens vs. OLMo1-1B at 2025B tokens trained. OLMo1-1B results are reported by evaluating AI2's checkpointed model (``step966000-tokens2025B'') with AI2's Open Language Model Evaluation System (OLMES, \citealt{gu2024olmes}). It is important to note that neither model is fully annealed at 2T tokens (Like OLMo1-1B, Gecko-1B also fully anneals between 3T and 4T tokens), however the comparison provides additional insight on training progress.

% \paragraph{\textsc{Gecko}-1B vs. OLMo1.}
% For a controlled comparison study, we configure the \textsc{Gecko}-1B model to closely follow the architectural hyperparameters of OLMo-1B (\citealt{Groeneveld2023OLMo}, July 2024 release). Both models have 16 blocks with a feature dimension of $d = 2048$. 
% We pretrained Gecko 1B with 2 trillion tokens on the same Dolma v1.7 dataset~\citep{dolma} that was used to train OLMo-1B.
% %ensuring that both models are trained on an identical corpus of approximately 1.7 trillion tokens. 
% We also use OLMo-1B's tokenizer with a vocabulary size of 50{,}304. Training is performed using the AdamW optimizer~\citep{loshchilov2019decoupled} with $\beta_1 = 0.9$, $\beta_2 = 0.95$, $\epsilon = 1\mathrm{e}{-8}$, and a peak learning rate of $4\mathrm{e}{-4}$. Cosine decay schedule is used, along with 2{,}000 warmup steps, weight decay of 0.05, gradient clipping of 1.0, and no dropout. 

\end{document}